# Optimal Placement and Patrolling of Autonomous Vehicles in Visibility-Based Robot Networks

Md Mahbubur Rahman[a], Leonardo Bobadilla[a], Franklin Abodo[a], Brian Rapp[b]

[a]*Florida International University, Miami, FL*
[b]*United States Army Research Lab, Aberdeen, MD*

## Abstract

In communication-denied or contested environments, Line-of-Sight (LoS) communication (e.g free space optical communication using infrared or visible light) becomes one of the most reliable and efficient ways to send information between geographically scattered mobile units. In this paper, we consider the problem of planning optimal locations and trajectories for a group of autonomous vehicles to see a set of units that are dispersed in an environment with obstacles. The contributions of the paper are the following: 1) We propose centralized and distributed algorithms to verify that the vehicles and units form a connected network through LoS; 2) We present an algorithm that can maintain visibility-based connectivity, if possible, by relocating a single vehicle; and 3) We study the computational complexity of calculating the optimal positioning of multiple vehicles and propose an approximate geometric procedure to restore visibility-based connected network. Our ideas are implemented and tested through realistic computer simulations and outdoor physical experiments.

## 1. Introduction

Communication between mobile units located in geographically separated positions in an environment is a matter of vital military importance. Communication can be easily interrupted by natural features such as terrain, atmospheric effects, and electromagnetic interference. Intentional jamming of communications and sniffing by an enemy may also pose a serious risk. In order to mitigate these problems, Line-of-Sight (LoS) communication can be established. This form of communication is more difficult to intercept or jam, because it requires the attacker to be directly between the sender and receiver. Because mission-related movements of land forces may naturally cause them to lose LoS with their friendly units, it is desirable to provide additional nodes or relays that can maintain communications between the units. A group of autonomous ground vehicles can fulfill this role, by moving from place to place as needed for the purposes of establishing relayed contact. One such environment is exemplified in Figure 1(a)-(b) where a LoS-based relay network has been established among the autonomous vehicles (rectangles) and units (circles).

Once a connected group of autonomous ground vehicles has been established in the field, its computational and storage capacity can be used to provide services to the units that it serves. This can provide additional military value, by analyzing tactical data, detecting threat patterns, or searching for information that would not otherwise be readily available. The task of a servicing ground vehicle is to maintain LoS to one or more units and at least one other servicing vehicle if more than one vehicle is deployed. Collectively they need to cover all the units and also to maintain a relay network among themselves.

As the units move around the environment, they frequently get isolated from the network as shown in Figure 1(c). This problem requires one or more autonomous vehicles to relocate into the visibility polygon of the disconnected unit in order to stay in the LoS and provide service. The relocation may damage the existing LoS-based connectivity of the network as shown in Figure 1(d) where the two vehicles become disconnected. Also, in some cases we may not have enough vehicles to form a fixed relay network.

This paper addresses the challenge of keeping units and vehicles connected so that no component gets isolated from the LoS network. Our proposed algorithms can identify a network disconnection when a movement occurs. Our main contributions are the analysis of the computational complexity of this class of problems and the proposal of different techniques that focus on the recovery of the relay network.

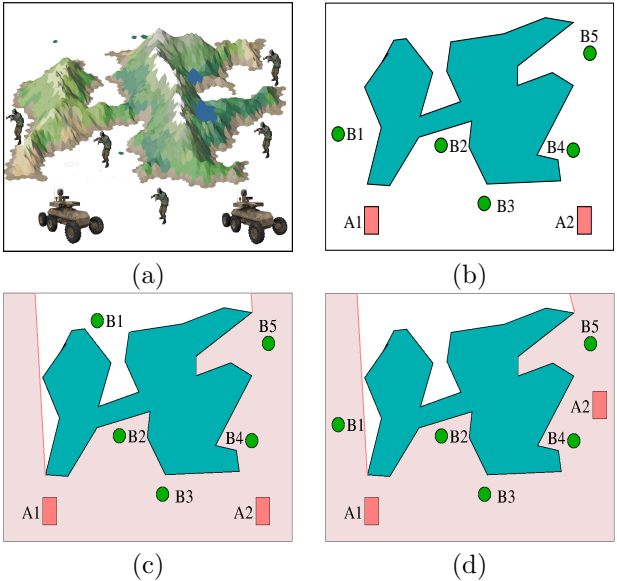

Figure 1: (a) A sample field mission where two autonomous servicing vehicles and five units are deployed; (b) The corresponding environment geometry in 2D space. The red rectangles are vehicles while the green circles are mobile units. The polygonal hole in the middle represents the obstacles and terrain $\mathcal{O}$; (c) A communication-invalid state where the unit $B_1$ is not seen by any of the vehicles; (d) Another communication-invalid state as vehicles $A_1$ and $A_2$ do not have any relay communication.

We evaluate our theories through extensive experiments on a realistic computer simulation model (Gazebo simulator [1]) and through outdoor hardware deployment. Our evaluation demonstrates the successful detection of network disconnection in various environments. In the event of disconnection, our proposed LoS based network recovery can compute new goal locations for the vehicles. We conduct an outdoor experiment where a vehicle equipped with a number of sensors (GPS, camera, ZigBee communication antennas) and onboard computation modules (Raspberry Pi, Arduino) is able to serve 2 units in an environment with obstacles.

*Contribution of this paper*: This paper is an extension to our previous work on visibility-based military unit formation [2]. We formulated the problem of validating whether the current configuration forms a valid connected network and proposed algorithms to compute a strategy to relocate a single vehicle to maintain LoS communication connectivity. Here, we extend this solution by developing a complete solution that relocates multiple vehicles to establish a relay network. We analyze the problem's computational complexity by relating it to well-known TSP [3] and set-cover [4] problems. We propose solutions using tools from Motion Planning and Computational Geometry to gain/regain a connected network by following a tour in order to maximize visibility. Finally, we implement our ideas and test them in various case studies using both software and hardware.

## 2. Related Work

Since the group of *servicing vehicles* must have LoS communication with all the units they serve, our ideas are naturally connected to *Art Gallery* problems [5, 6] and other visibility-based approaches in computational geometry [7]. Art-gallery based approaches have been used in robotics to solve sensor [8] and landmark placement [9] problems. Another computational geometry problem that is connected to our ideas is the *Watchman Route* problem [10]. Some of the differences between the traditional Watchman Route problem and our setup are: 1) We are not only concerned about the shortest path but also about a path that will keep the most visibility with all units and other vehicles; and 2) The vehicle's paths should respect differential constraints.

In [11], the authors proposed a scheme to visit all the visibility polygons using a single vehicle, which leads to redundancy when polygons intersect. Also, they do not consider formation of a constrained relay network among *multiple* robots, which is the main goal of this work. The solution is based on a genetic algorithm, and the patrolling route computed for a single robot is unable to guarantee the optimality due to random mutation. In contrast, here we propose a solution that minimizes the number of regions to visit by a single vehicle that either eliminates or reduces the patrolling route.

Closely related to our problem are *visibility-based pursuit* schemes whose goal is to find a path that will guarantee that an evader is captured regardless of his motion [12]. Our work is also closely connected to path planning approaches that attempt to maintain visibility to a single static landmark [13, 14].

Another stream of research deliberately allow the robots to lose connectivity among themselves while doing cooperative surveillance in a region [15, 16, 17, 18]. These researches do not incorporate the visibility based connectivity which is by nature more constrained. Additionally, we do not assume to lose connectivity at the beginning while solving the problem. Rather our approach greedily tries to make the robots as much connected and static as possible unless a disconnection is required.

The idea of a powerful mobile unit uploading, downloading, and distributing data to a set of dynamic units has been explored in *data muling* and *data ferrying* [19, 20, 21, 22]). One important difference between our formulation and the data muling approach is that communication is based on Line-of-Sight instead of the proximity of the sensor nodes. In the area of communication, Free-Space Optical Communications (FSOC) [23] is being considered as an alternative for military network-centric operations. Particularly related is the work in [24] where the problem of two mobile nodes that try to maintain LoS alignment is studied.

# 3. Problem Statement

Let $A_1, A_2, \ldots, A_n$ be a set of $n$ servicing *vehicles*, with configuration spaces $\mathcal{C}_1, \mathcal{C}_2, \ldots, \mathcal{C}_n$ and a set of $m$ mobile *units* be, $B_1, B_2, \ldots, B_m$. Units and vehicles are deployed in a 2D world $\mathcal{W} = \mathbb{R}^2$ which we assume to be a connected polygon. Let $\mathcal{O}$ be the set of obstacles that block communication and are modeled as polygons. The collision-free space is defined as $E = \mathcal{W} \setminus \mathcal{O}$. The mobile units can move freely in the world and are modeled as point robots without rotation. Accordingly, the configuration for a mobile unit $B_j$ is defined as, $r_j = (x, y) \in E$, and the set $\mathcal{B} = E^m$ is the configuration space for the mobile units. The servicing vehicles can move inside the bounded environment $E$ using both translation and rotation actions, and are modeled as point robots with an orientation and configuration defined as $q_i = (x, y, \theta) \in E \times [0, 2\pi)$ [25]. These vehicles are car-like, and a given vehicle $A_i$ must satisfy differential constraints and dynamics defined as $\dot{x}_i = u_s^i \cos\theta; \dot{y}_i = u_s^i \sin\theta$, and $\dot{\theta}_i = \frac{u_s^i}{L^i} \tan u_\phi^i$; [26], where $u_s^i$ is the forward speed and $u_\phi^i$ is the steering angle of the vehicle. Together, the $n$ vehicles compose the configuration space $\mathcal{C} = \mathcal{C}_1 \times \mathcal{C}_2 \times \cdots \times \mathcal{C}_n$. We define the entire system state space to be $X = \mathcal{C} \times \mathcal{B}$. Let $X_{obs} = \{x \in X : x \cap O \neq \emptyset \text{ where } O \in \mathcal{O}\}$ be the obstacle state space. The collision-free state space is then $X_{free} = X \setminus X_{obs}$.

## 3.1. Communication State Validity

Communication can only be established among servicing vehicles and between servicing vehicles and mobile units through LoS. Mobile units cannot communicate with each other. The communication range can be characterized by a *visibility polygon*. The visibility polygon $V(p)$ for a point $p \in E$ is defined as [27]:

$$V(p) = \{w | w \in E \text{ and } \overline{pw} \cap E = \overline{pw}\}. \quad (1)$$

**Definition 3.1.** *A state $x \in X$ is considered* communication-valid *if and only if each unit is visible by at least one servicing vehicle and the servicing vehicles form a connected network. We define this set of configurations as $X_{comm} \subset X$.*

According to the definition, we must satisfy the following conditions in order to have a communication-valid state:

$$\forall j, \exists i \text{ s.t. } r_j \in V(q_i) \text{ for } 1 \leq j \leq m \text{ and } 1 \leq i \leq n \quad (2)$$

$$\{(q_j, q_k) | q_k \in V(q_j) \text{ for } 1 \leq j, k \leq n, \ k \neq j\} \equiv CC(x) \quad (3)$$

where $CC(x)$ is a *connected component* formed by all the vehicle-vehicle connections.

The state $x \in X$ is changed whenever a unit or vehicle changes its configuration. The units move autonomously, and in response to those movements we may need to plan the trajectories and new configurations for the servicing vehicles depending on communication-validity. Therefore, we have a decision problem in which we want to know whether a given state is *communication-valid* ($x \in X_{comm}$). This problem can be formulated as follows:

**Problem 1: Communication State Validation**
*Given the workspace $\mathcal{W}$, a set of obstacles $\mathcal{O}$, a set of configurations $\mathcal{C}$ for servicing vehicles, and $\mathcal{B}$ for mobile units, determine whether a state $x \in X_{comm}$ or not.*

## 3.2. Invalid-to-Valid Communication State Restoration

Initially, we assume that the visibility-based network is connected as shown in Figure 1 and $x \in X_{comm}$. Since the mobile units are allowed to move freely throughout the environment $E$, the system becomes *communication-invalid* frequently. A unit is marked as disconnected if and only if it is not visible to any of the servicing vehicles. A set of disconnected units, $D \subseteq \{B_1, B_2, \ldots, B_m\}$, is defined based on a given state $x \in X$, and we dispatch an available vehicle $A_i$ from its current location $x_s^i$ to a newly computed goal region $X_G^i$ in order to reconnect the strayed units. As the event of vehicle relocation must not break the existing partially connected network, the selection of vehicles to be moved must be done carefully. This motivates the following problem:

**Problem 2: Communication Validity Restoration**
*Given $\mathcal{W}$ and $\mathcal{O}$, the current state $x \in X$, and a set of disconnected units $D$, select one or a number of vehicles to relocate and compute their new goal regions, $X_G$, that will reconnect all the units in $D$.*

## 3.3. Patrolling and Trajectory Estimation

There may be situations where there are not enough vehicles to serve all of the units and maintain a connected relay network. In these cases, we need to calculate the optimal regions on the free space so that placing the available vehicles on those areas can serve as many units as possible. Furthermore, a patrolling tour may be required by one vehicle which will connect the remaining disconnected units and vehicles and act like a dynamic relay link. Because the vehicle designated to serve the disconnected units and other vehicles along the patrolling route may lose its existing connection to one or more units or vehicles that it is already servicing, the tour must be chosen optimally, not arbitrarily, such that the traveling time is minimized. Therefore, if we have a set of disconnected units, $D$, a tour, $\tau : [0, T] \to X_{free}$, must satisfy,

$$\forall i \exists t \text{ s.t. } r_i \in V(\tau(t)) \text{ where } 1 \leq i \leq m, t \in [0, T]. \quad (4)$$

**Problem 3: Patrolling Trajectory Estimation**
*Given $\mathcal{W}$ and $\mathcal{O}$, the current state $x \in X$ and a set of disconnected units $D$, compute the optimal patrolling trajectory $\tau : [0, T] \to X_{free}$, such that $\tau$ touches the minimum number of discrete regions.*

# 4. Communication State Validation

## 4.1. Centralized Algorithm

In this section we provide the solution to verify whether the current state $x \in X$ is communication-valid, which solves *Problem 1* demonstrated in Section 3.1. Initially, we are given the positions of units and vehicles and we do not know which vehicle is providing service to which unit. In order to solve *Problem 1*, we propose Algorithm 1 that works based on graph theoretic network connectivity [28, 29] solutions. A visibility-based graph can be constructed where the node set is composed of all the components (vehicles and units) and an edge is added between two nodes if the corresponding components are visible to each other. However, checking the algebraic connectivity [28] on this graph is not sufficient. For example, the graph shown in Figure 2(b) is connected but not communication-valid. This type of graph occurs if there are obstacles between vehicles. Therefore our proposed validation is two-fold and we generate the following two types of undirected graphs based on a particular state $x$.

*Vehicle Relay Graph ($\mathcal{G}_\mathcal{A}$):* This state-dependent undirected graph is a mapping $g_\mathcal{A} : X \to \mathcal{G}_\mathcal{A}(\mathcal{V}_\mathcal{A}, \mathcal{E}_\mathcal{A})$, where $\mathcal{V}_\mathcal{A} = \{A_1, A_2, \ldots, A_n\}$ is the set of vehicle nodes (see Figure 2(a)). $\mathcal{E}_\mathcal{A}$ denotes the set of edges defined as,

$$\mathcal{E}_\mathcal{A} = \{e_{ij} | q_i \in V(q_j)\} \quad (5)$$

where $q_i$ and $q_j$ are the positions of vehicles $A_i$ and $A_j$ in the environment $E$. This implies that an edge $e_{ij}$ exists if and only if the vehicle $A_i$ is inside the visibility polygon, $V(q_j)$, of vehicle $A_j$. We then compute the $n \times n$ Laplacian matrix, $\mathcal{L}(\mathcal{G}_\mathcal{A}) = DEG(G_\mathcal{A}) - ADJ(\mathcal{G}_\mathcal{A})$, where $ADG(\mathcal{G}_\mathcal{A})$ is the familiar $(0, 1)$ adjacency matrix, and $DEG(\mathcal{G}_\mathcal{A})$ is the diagonal matrix of vertex degrees [30], also called the valency matrix of $\mathcal{G}_\mathcal{A}$. The entries of $\mathcal{L}$ are as follows [31]:

i) $l_{ij} = \begin{cases} -1 & \text{if an edge exists between } i \text{ and } j \\ 0 & \text{otherwise} \end{cases}$

ii) $l_{ii} = -\sum_{k=1, k \neq i}^{n} l_{ik}$

In line 2 of Algorithm 1, we check the second-smallest eigenvalue $\lambda_2(\mathcal{L}(\mathcal{G}_\mathcal{A}))$ of $\mathcal{G}_\mathcal{A}$ to see whether it is positive. A non-positive value indicates that the relay network formed by all the vehicles does not exist and the network is communication-invalid (line 3). If $\lambda_2 > 0$, then we go to the second step of validation where we check the entire network connectivity (lines $5 - 11$).

*Unit Graph ($\mathcal{G}_\mathcal{B}$):* The unit graph $\mathcal{G}_\mathcal{B}$, which is also undirected and is a mapping $g_\mathcal{B} : X \to \mathcal{G}_\mathcal{B}(\mathcal{V}_\mathcal{B}, \mathcal{E}_\mathcal{B})$, is computed in line 5. In contrast to $G_\mathcal{A}$, the graph $\mathcal{G}_\mathcal{B}$ includes all the $m$ units and $n$ vehicles in its node set $\mathcal{V}_\mathcal{B}$. Accordingly, $\mathcal{V}_\mathcal{B} = \{A_1, A_2, \ldots, A_n, B_1, B_2, \ldots, B_m\}$ is indexed by the vehicles, followed by the units, so that all units have indices greater than $n$. The edge set $\mathcal{E}_\mathcal{B}$ has the following form:

$$\mathcal{E}_\mathcal{B} = \{e_{ij} | r_j \in V(q_i) \text{ where } n < j \leq n+m \text{ and } 0 < i \leq n\}. \quad (6)$$

This means that an edge is added if and only if a unit's position $r_j$ is visible from some vehicle's position $q_i$ (see Figure 2(b)).

Finally, we form a state-dependent graph $\mathcal{G}(\mathcal{V}, \mathcal{E})$ as shown in Figure 2(c), which is the union of the two graphs $\mathcal{G}_\mathcal{A}$ and $\mathcal{G}_\mathcal{B}$. Accordingly, the vertex set $\mathcal{V} = \mathcal{V}_\mathcal{A} \cup \mathcal{V}_\mathcal{B}$ and the edge set $\mathcal{E} = \mathcal{E}_\mathcal{A} \cup \mathcal{E}_\mathcal{B}$. Therefore we conclude that the graph is communication-valid if the second-smallest eigenvalue $\lambda_2$ of the $(m+n) \times (m+n)$ Laplacian matrix, $\mathcal{L}(\mathcal{G})$, of graph $\mathcal{G}(\mathcal{V}, \mathcal{E})$ is greater than *zero* (lines $7 - 11$).

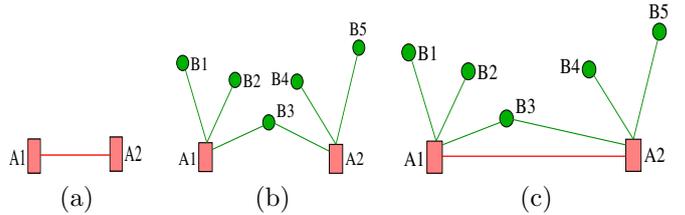

Figure 2: (a) Vehicle relay graph $\mathcal{G}_\mathcal{A}$ generated from a vehicle-vehicle relay network for the environment demonstrated in Figure 1; (b) Unit graph $\mathcal{G}_\mathcal{B}$ from the vehicle-unit connectivity; (c) Union of the two graphs, $\mathcal{G} = \mathcal{G}_\mathcal{A} \cup \mathcal{G}_\mathcal{B}$.

---

**Algorithm 1** communicationCheck($x, \mathcal{O}$)

1: $\mathcal{G}_\mathcal{A} = g_\mathcal{A}(x)$
2: **if** $\lambda_2(\mathcal{L}(\mathcal{G}_\mathcal{A})) \leq 0$ **then**
3:     *return false*
4: **end if**
5: $\mathcal{G}_\mathcal{B} = g_\mathcal{B}(x)$
6: $\mathcal{G} = \mathcal{G}_\mathcal{A} \cup \mathcal{G}_\mathcal{B}$
7: **if** $\lambda_2(\mathcal{L}(\mathcal{G})) \leq 0$ **then**
8:     *return false*
9: **else**
10:     *return true*
11: **end if**

---

*Analysis of Algorithm 1*: The graph creation in lines 1 and 5 uses a visibility polygon computation algorithm to determine the edges of the graphs. Each polygon computation takes $O(n)$ [27] and for $n$ vehicles the running time is $O(n^2)$. The dominant factor, however, is in computing the eigenvalues (lines 2 and 7) which generally takes $O(n^3)$ in the worst case. Therefore the running time of Algorithm 1 is $O(n^3)$.

## 4.2. Distributed Algorithm

An improvement can be made over the centralized Algorithm 1 if we use the computational power of all the vehicles in a distributed manner. Algorithm 2 presents pseudo code for the *distributed* communication-validity checking *program* which will run in each of the vehicles. In a *distributed processing* system we rely on message passing through network protocols. Algorithm 2 will be triggered once it receives a request or control message. In Line 1 we collect all the elements visible from the vehicle. Line 2 sends a query message to each of the neighboring vehicles except the requester ($reqV$), to share their coverage information. The program then waits for all the vehicles' response messages. Accordingly, line 4 merges the vehicle's own visibility information with that of its neighbors. If the current vehicle is the *initiator*, lines $5-10$ will check the network to see if all the units and vehicles in service were discovered or not. Otherwise, in lines $11-13$, the resulting status from the current vehicle will be sent back to the requester.

---

**Algorithm 2** stableStateDist($C_i, reqV$)
---
1: $\alpha = h_i(x)$
2: $query\ h_j(x)\ to\ all\ adjacent\ vehicles, A_j \in N \setminus reqV$
3: $wait\ and\ receive\ h_j(x)\ from\ all\ neighbor\ vehicles\ A_j$

4: $\alpha = \alpha \cup h_1(1) \cup h_2(x) \cdots \cup h_j(x)$
5: **if** $A_i\ is\ initiator()$ **then**
6:   **if** $\alpha == N \cup M$ **then**
7:     $return\ true$
8:   **end if**
9:   $return\ false$
10: **end if**
11: **if** $receiver(A_i)$ **then**
12:   $send\ \alpha\ to\ the\ requester$
13: **end if**

---

*4.2.1. Analysis of Algorithm 2*

All the lines in Algorithm 2 except lines $2, 3$ and $12$ run in $O(1)$. Lines 2 and 3 will run at most $O(n)$ if all other $n-1$ vehicles are visible. Therefore on a single vehicle the algorithm takes $O(n)$ time. Similarly if all $n-1$ other vehicles are the requester, line 12 will take $O(n)$. The drawback of this algorithm is the *messaging* overhead and *waiting time* for responses. The number of messages being sent can be vast if the graph is *dense*.

## 5. Recovering a Communication-valid State with a Single Vehicle

As units are on the move, this may result in disconnections from their respective servicing vehicles. Here we propose a solution that dispatches a single vehicle in order to reconnect a strayed unit from the visibility-based network. Any movement inside a network triggers Algorithm 3, which identifies any disconnections and relocates a vehicle that best re-establishes a communication-valid state without affecting existing network connections. The set of disconnected units $D$ is defined as:

$$D = \{B_j | \forall i, r_j \notin V(q_i)\ \text{where}\ 1 \leq i \leq n\}. \tag{7}$$

Therefore $D$ is the set of units that are not visible to any of the vehicles due to obstacles. In other words, the set of all the units with degree *zero* in the graph $\mathcal{G}_\mathcal{B}$ compose the disconnected set $D$. If there is any such non-visible unit (i,e., $D \neq \emptyset$), we attempt to resolve disconnections for each $B_j \in D$ using Algorithm 3.

Next, we define the set $H_i$ as the set of *hard constrained* units of a vehicle $A_i$ which are only visible from $A_i$ and to which no other vehicles can provide service. A unit is said to be a hard constrained unit if and only if it is visible from only one vehicle. Lines $1-3$ of Algorithm 3 compute $H_i$ for all the vehicles according to the following equation:

$$H_i = \{r_j | r_j \in V(q_i)\ \text{and}\ \forall k \neq i, r_j \notin V(q_k)\}. \tag{8}$$

Therefore, the unit nodes (nodes corresponding to the units) with degree *one* in the graph $\mathcal{G}_\mathcal{B}$ are the members of the hard constrained sets. In line 4 we compute the intersecting polygon $V(H_i)$ of all visibility polygons of all members in $H_i$. Initially the set of candidate vehicles for relocation is $C = \{A_1, A_2, \ldots, A_n\}$. However, we may not be able to relocate all the vehicles in $C$ as this may break the existing connected graph topology $\mathcal{G}_\mathcal{A}$ among the vehicles. Therefore, we check the second-smallest eigenvalue of the Laplacian matrix of a graph generated by removing the corresponding vehicle nodes $A_i \in C$ along with their incident edges from graph $\mathcal{G}_\mathcal{A}$. We remove the nodes from $C$ that make $\lambda_2 \leq 0$ (line 6 of Algorithm 3).

The new goal polygon $X_G^i$ of a candidate vehicle $A_i \in C$ must be inside the visibility polygons of 1) the disconnected unit $B_j$ and 2) at least one other vehicle that is a part of the existing relay network. Moreover, if there is any hard constrained unit and $H_i \neq \emptyset$ then $X_G^i$ must be inside the polygon $V(H_i)$. As the visibility polygons may be concave in an environment filled with obstacles, we may get multiple goal polygons. In such cases, we take the largest one. Therefore we compute $X_G^i$ for $A_i \in C$ as follows (see line 8):

$$X_G^i = \begin{cases} \max\limits_{A_k \neq A_i, 1 \leq k \leq n} V(r_j) \cap V(q_k); & \text{if } H_i = \emptyset \\ \max\limits_{A_k \neq A_i, 1 \leq k \leq n} V(r_j) \cap V(q_k) \cap V(H_i); & \text{otherwise} \end{cases} \tag{9}$$

Once the goal regions for all the candidate vehicles in $C$ are computed, we only retain the vehicles that have nonempty goal regions (line 10). We then select the optimal vehicle $A_{r_j}$ in terms of the motion cost. In brief, the $motionCost()$ method in line 14 computes the relocation cost of a vehicle from its current position $x_s^i$ to the computed goal region $X_G^i$ using a motion planning algorithm such as Rapidly-exploring Random Trees Star (RRT*) [32].

**Algorithm 3** singleMoveComm ($B_j \in D, \mathcal{G_B}, \mathcal{G_A}$)

1: **for** each vehicle $A_i$ **do**
2:     $H_i = computeHardConstrained(\mathcal{G_B})$
3: **end for**
4: $V(H_i) = \bigcap_{k=1}^{|H_i|} V(r_k \in H_i)$
5: $C = \{A_i | 1 \leq i \leq n\}$
6: $C = C \setminus A_i$ s.t. $\lambda_2(\mathcal{L}(\mathcal{G_A}(\mathcal{V_A} \setminus A_i, \mathcal{E_A} \setminus e_i))) \leq 0$
7: **for** $A_i \in C$ **do**
8:     $X_G^i = \begin{cases} \max_{A_k \neq A_i, 1 \leq k \leq n} V(r_j) \cap V(q_k); & \text{if } V(H_i) = \emptyset \\ \max_{A_k \neq A_i, 1 \leq k \leq n} V(r_j) \cap V(q_k) \cap V(H_i); & \text{otherwise} \end{cases}$
9: **end for**
10: $C = C \setminus A_i$ s.t. $X_G^i = \emptyset$
11: **if** $X_G^i = \emptyset$ for all $1 \leq i \leq |C|$ **then**
12:     *return failure*
13: **end if**
14: $A_{r_j} = \underset{A_i \in C}{\operatorname{argmin}}[motionCost(x_s^i, X_G^i)]$
15: *return success*

## 6. Recovering a Communication-valid State with Multiple Vehicles

### 6.1. Hardness of Relocating Multiple Robots

If *Problem 2* cannot be solved by Algorithm 3, we need to move more than one vehicle. New configurations for more than one vehicle cannot be calculated efficiently as different combinations of vehicle movements are possible. Another important constraint is the number of vehicles; a sufficient number of vehicles may not be available to support all the units. Therefore, we first assume that we have only one vehicle that follows a travelling route $\tau$ ([11]) to visit the visibility polygons of the $m$ units. An analysis of the hardness of this problem follows.

*Definition 6.0: LoS Communication Problem :* Given a set of $m$ visibility polygons each for one unit $B_i \in M$, find the shortest tour $\tau : [0, T] \to X_{free}$ to visit at least one point in each polygon.

**Proposition 6.1.** *LoS Communication problem is NP-Hard.*

PROOF. We will prove the hardness of the problem by polynomially reducing the *Traveling Salesman Problem with Neighbors (TSPN)* [3], a well-known NP-hard problem, to our LoS communication problem. Suppose TSPN takes as an input a set of convex polygons, $\Gamma = \{P_1, P_2, \ldots, P_m\}$, and the goal is to find a minimum cost tour that touches at least one point in each of the polygons. The convexity of the polygon does not reduce the difficulty of the problem [3].

The reduction algorithm will take as an input $\Gamma$ from TSPN and will place a unit, $B_i$, in the centroid of each polygon $P_i$ (see Figure 3). This centroid calculation can be done in polynomial time. Therefore, $P_i$ will work as visibility polygon for the unit $B_i$. These will convert the input of TSPN problem to the input of LoS Communication in polynomial time. The output transformation is trivial, since the solution for the LoS Communication problem $\tau$ is clearly a solution for the TSPN as each of the units $B_i \in M$ belong to a polygon $P_i$ that will be touched by the tour, solving the TSPN.

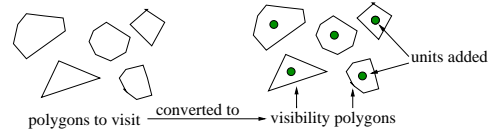

Figure 3: An instance of TSPN is reduced to an instance of LoS Communication problem. Each polygon in TSPN will act as visibility polygon of an assigned unit.

Conversely, suppose that we have a solution tour, $\tau$ for the TSPN that touches each of the polygons $P_i \in \Gamma$. We can use this as a route that will be followed by the vehicle in our LoS communication problem since the vehicle will go into all the visibility polygons in each tour and will provide service to all the units.

### 6.2. Approximated Solution

Since our problem is NP-hard, it cannot be solved exactly in polynomial time unless P=NP. We must use an approximation algorithm for TSPN to get a near optimal solution. As previously mentioned in Section 2, a genetic algorithm solution of this problem can be found in [11] for a UAV where the aerial robot is allowed to fly over the obstacles. However, we cannot use this solution or the approximate solution for TSPN due to obstacles in the environment. We must instead use a motion planning algorithm once the sequence of polygons to visit is computed.

Since visibility polygons may intersect, the number of regions to visit may be fewer than the total number of individual polygons; this allows for a significant improvement over the works in the literature ([11]). Let $\Gamma$ be the set of goal regions to be visited that are inside the area composed by the visibility polygons of all the units ($\Gamma \subseteq \bigcup_{i=1}^{m} V(B_i)$). There is an intractable number of regions and sub regions in the obstacle free plane $E$. Therefore, we developed Algorithm 4 which computes $\Gamma$, the finite set of goal polygons to be visited by a vehicle. In line 1, we compute the visibility polygons $V(B_1), V(B_2), \ldots, V(B_m)$ corresponding to the units. Next, we decompose the obstacle free environment $E$ into a countable set of polygonal faces $\mathcal{F} = \{P_1, P_2, \ldots, P_\rho\}$ based on the intersections of the visibility polygons as shown in Figure 4. We need to select a set of polygons $\Gamma$ from $\mathcal{F}$, which implies that $\Gamma \subseteq \mathcal{F}$. One important fact is that $\mathcal{F}$ contains all the original polygons and the split polygons resulting from their intersections after the decomposition process. This helps us to select bigger polygonal regions for vehicle placements that cover large areas, making it easier for the motion planner to compute paths, given large goal regions.

The visibility-based decomposition is a vital step towards solving the problem as the edges of a face $P \in \mathcal{F}$ inflict at least one visibility event. Crossing an edge results in the appearance or disappearance of a unit. As a result, each of the polygonal faces $P$ is visible by a set of units. We need to choose the minimum number of such polygons so that they collectively cover all the units. This resembles the well-known geometric *Set Cover* problem [33, 34], and computing the optimal solution is NP-Hard. An instance of such problem consists of a finite set $\mathcal{U} = \{B_1, B_2, \ldots, B_m\}$ and a set of allowable polygons $\mathcal{F} = \{P_1, P_2, \ldots, P_\rho\}$, such that every point in $\mathcal{U}$ is covered by at least one polygon in $\mathcal{F}$ [33]. We use a modified greedy set cover approximation algorithm [35, 4] to solve this problem (lines 3-11 of algorithm 4).

In line 3 of algorithm 4, we assign a label $y_P$ to a polygon $P$, which is a set,

$$y_P = \{B_j, B_k, \ldots, B_l\}; \forall c \in \{j, k, \ldots, l\} \ V(B_c) \cap P_i = P_i. \tag{10}$$

This means the label $y_P$ of a polygon $P$ contains the names of the units whose visibility polygons completely enclose $P$. Next we assign a score to all the polygons as,

$$\hat{s}(P) = \gamma \cdot area(P) + \sum_{B_k \in y_p} \left[ \alpha - \beta \cdot d(P, B_k) \right] \tag{11}$$

Here, $area : \mathcal{F} \to \mathbb{R}^{\geq 0}$ is used to compute the area of the polygon and $d : \mathcal{F} \times \mathcal{B} \to \mathbb{R}^{\geq 0}$ is the distance function to compute the distance between any visible unit and the polygon. $\alpha, \beta, \gamma \in \mathbb{R}^{\geq 0}$ are the variables and $\alpha$ is chosen to be very large compared to $\beta$ and $\gamma$ to make the visible number of units ($|y_P|$) of the polygon $P$ the most dominant factor of $\hat{s}$. Therefore, it is obvious that the highest scoring polygon covers most of the units. In the cases where more that one polygons cover same number of units, we chose the largest and the nearest one to the units.

Line 7 of Algorithm 4 greedily selects a polygon that covers as many units as possible. Breaking the tie is done using the score $\hat{s}$. We remove the units from $\mathcal{U}$ in line 8 that are covered by $P$ (i.e. in $y_P$) and add the polygon in the resulting set $\Gamma$ in line 9. This process ends when $\mathcal{U}$ becomes empty and we terminate the algorithm by returning $\Gamma$ as the goal regions to visit.

*Analysis*: Algorithm 4 is composed of two different algorithms. Lines 1-4 calculate a set of polygons $\mathcal{F}$ which is fed as an input to the set cover approximation of lines 5-11. The polygon set $\mathcal{F}$ is produced through the intersection of $m$ visibility polygons which are concave. From a pairwise visibility polygon intersection, we get $O(cm^2)$ polygons for some constant $c$. These resultant polygons also intersect, yielding $O(c^2 m^4)$ polygons. This implies $|\mathcal{F}| = c^2 m^4$, which is the input of the set cover approximation that runs in $O(|\mathcal{U}||\mathcal{F}| \min(|\mathcal{U}|, |\mathcal{F}|))$ time [35].

**Algorithm 4** multiRobotPlacement ($\mathcal{B}, \mathcal{O}$)
1: $\mathbb{V} = \{V(B_1), V(B_2), \ldots, V(B_m)\}$
2: $\mathcal{F} = decompose(\mathbb{V})$
3: $\forall P \in \mathcal{F}, \ y_P = assignLabel(\mathbb{V})$
4: $\forall P \in \mathcal{F}, \ \hat{s}(t_i) = assignScore(P, y_P)$
5: $\Gamma = \emptyset; \ \mathcal{U} = \{B_1, B_2, \ldots, B_m\}$
6: **while** $\mathcal{U} \neq \emptyset$ **do**
7: $\quad$ Select $P \in \mathcal{F}$ that maximizes $|y_P \cap \mathcal{U}|$
8: $\quad \mathcal{U} = \mathcal{U} - y_P$
9: $\quad \Gamma = \Gamma \cup \{P\}$
10: **end while**
11: *return* $\Gamma$

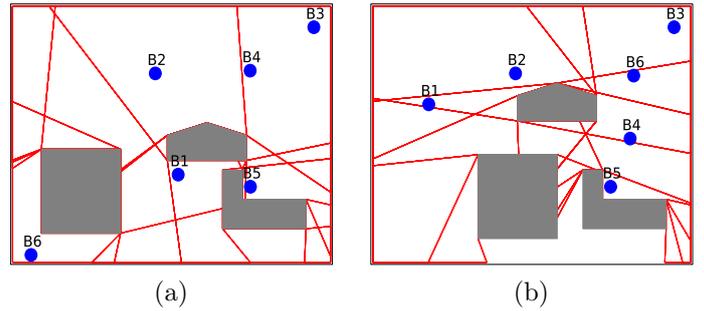

Figure 4: Two sample environments are partitioned using visibility polygon based decomposition.

### 6.3. Multi-Robot Placements and Patrolling

In the presence of $n$ vehicles, we have three different cases:

Case $|\Gamma| = 1$: This is a trivial case where we deploy a vehicle in the sole polygon that is visible to all the units.

Case $|\Gamma| < n$: In this case we have enough vehicles that they can be placed as static servers to each of the polygons. We use $|\Gamma|$ vehicles where each of the polygons receives one vehicle. We may need to use one vehicle for patrolling among the polygons depending on the components in the visibility-based vehicle graph ($\mathcal{G}_\mathcal{A}$).

Case $|\Gamma| \geq n$ and $n > 1$: In such cases we do not have sufficient vehicles to cover all the polygons. Therefore, we keep assigning one vehicle per polygon in $\Gamma$, prioritizing based on their scores, $\hat{s}$, until we are left with one vehicle. The last vehicle will perform a tour, $\tau$, among all the polygons. Therefore the polygons $P_1, P_2, \ldots P_{n-1}$ are covered and all other polygons $P_i \in \Gamma$ s.t. $i \geq n$ are not covered.

In both of the above cases, we may need an optimal patrolling strategy in order to establish a dynamic link among the covered polygons (having an assigned vehicles) and uncovered polygons (having no assigned vehicle). Once a number of available vehicles are deployed, as shown in Figure 5(a), we compute the vehicle-graph $\mathcal{G}_\mathcal{A}(\mathcal{V}_\mathcal{A}, \mathcal{E}_\mathcal{A})$ as explained earlier in Section 4.1 (see Figure 5(b)). We then apply the connected component algorithm [36] to get a set of subgraph components $C_1, C_2, \ldots, C_\kappa$ where $\mathcal{V}_\mathcal{A} = \bigcup_{i=1}^{\kappa} C_i$. By definition, any two member vertices

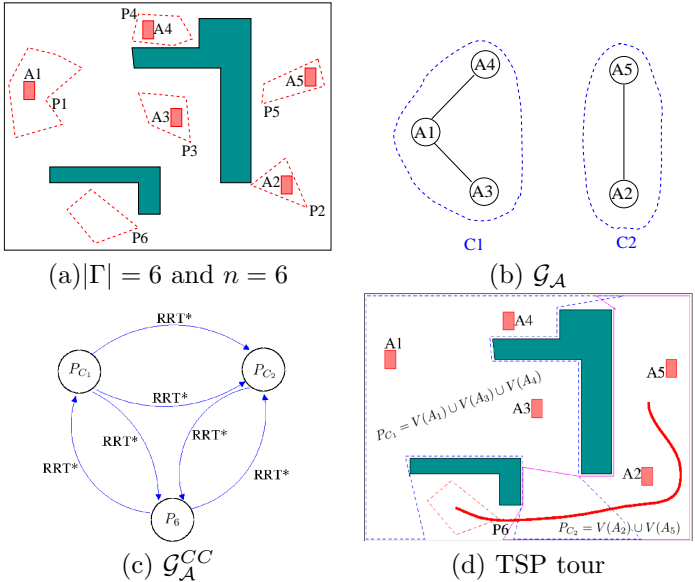
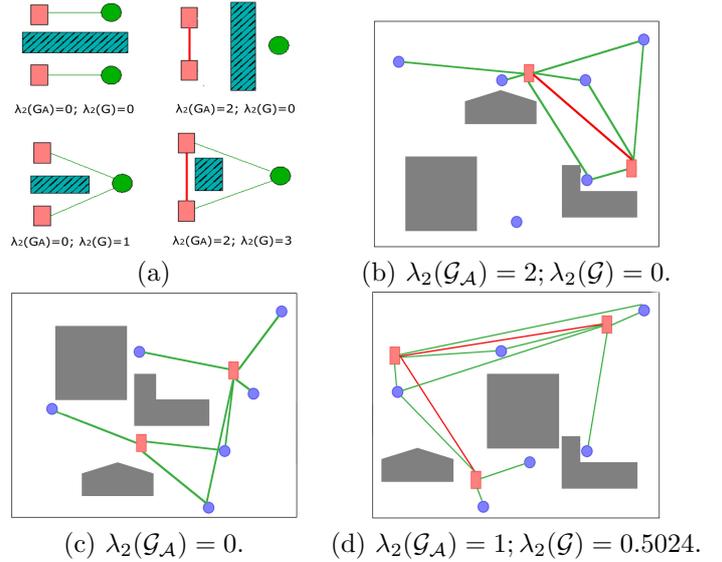

Figure 5: (a) A set of six polygons, $\Gamma = \{P_1, P_2, P_3, P_4, P_5, P_6\}$ computed by approximate set cover (Algorithm 4) that are to be covered by $n = 6$ available vehicles; (b) Two connected components $C_1$ and $C_2$ are computed from vehicle graph $\mathcal{G}_\mathcal{A}$; (c) Connected component graph $\mathcal{G}_\mathcal{A}^{CC}$; (d) TSP tour and RRT* path to be followed by the 6-th vehicle.

Figure 6: (a) A few trivial environment setups. Only the bottom right state is communication-valid; (b) and (c) are two communication-invalid states as $\lambda_2 \leq 0$ for at least one graph (relay or union graph) in each environment. (d) A communication-valid state as $\lambda_2(\mathcal{G}_\mathcal{A}) > 0$ and $\lambda_2(\mathcal{G}) > 0$.

$A_j, A_k$ in a component $C_i$ are connected through a path (visibility path in our case) as shown in Figure 5(b). We merge all the visibility polygons of the vehicles under a component $C_i$ to make a single polygon,

$$P_{C_i} = \bigcup_{A_j \in C_i} V(A_j) \qquad (12)$$

There may be some polygons that are not covered due to insufficient vehicles (if $|\Gamma| \geq n$). These are the polygons denoted as $\Gamma^U = \Gamma \setminus \{P_1, \ldots, P_{|\mathcal{V}_\mathcal{A}|}\}$. We thereafter create a directed connected-component graph $\mathcal{G}_\mathcal{A}^{CC}(\mathcal{V}_\mathcal{A}^{CC}, \mathcal{E}_\mathcal{A}^{CC})$, as shown in Figure 5(c), where the vertices are composed of the component polygons and uncovered polygons,

$$\mathcal{V}_\mathcal{A}^{CC} = \{P_{C_1}, P_{C_2}, \ldots, P_{C_\kappa}\} \cup \Gamma^U \qquad (13)$$

Graph $\mathcal{G}_\mathcal{A}^{CC}$ is a complete graph which means any polygon is reachable from any other polygon, as our environment $E$ is connected. The weighted directed edge $e_{ij}^{CC} \in \mathcal{E}_\mathcal{A}^{CC}$ between two vertices $P_i^{CC}, P_j^{CC} \in \mathcal{V}_\mathcal{A}^{CC}$ is computed using a motion planning algorithm such as RRT*, A* or combinatorial planning [32, 25], that finds a path (edge) between the polygons while avoiding the set of obstacles $\mathcal{O}$. Once all the edges are computed, we apply the approximate Geometric TSP [37] algorithm to compute the sequence of polygons to visit. Finally, a motion planner computes a sub-optimal tour that touches all the polygons according to the sequence with minimal motion cost (see Figure 5(d)).

## 7. Experimental Results

### 7.1. Checking Communication-Valid State

In the first case study, we validate the correctness of Algorithm 1 to check the communication-valid state space. The results from our experiments on different setups of the environment are shown in Figure 6. In Figure 6(a) we have a few trivial environments where only one graph is communication-valid (bottom right with $\lambda_2(\mathcal{G}_\mathcal{A}) = 2$ and $\lambda_2(\mathcal{G}) = 3$). A complex environment with three obstacles, two vehicles and six units is presented in Figure 6(b). Here, the relay network is connected, as the second-smallest eigenvalue of the vehicle graph's Laplacian is $\lambda_2(\mathcal{G}_\mathcal{A}) = 2 > 0$. However, the union graph including vehicles and units results in $\lambda_2(\mathcal{G}) = 0$, which indicates that the setup is not communication-valid.

Another environment is shown in Figure 6(c) where the relay network is not communication-valid ($\lambda_2(\mathcal{G}_\mathcal{A}) = 0$), although the union graph is connected. Finally, in Figure 6(d) we demonstrate a communication-valid network with three vehicles where both the relay graph and union graph are connected ($\lambda_2(\mathcal{G}_\mathcal{A}) = 1$ and $\lambda_2(\mathcal{G}) = 0.5024$).

### 7.2. Regaining a Communication-valid State by Single Vehicle Movement

We used the Bonnmotion Library [38] to generate different mobility models. The CGAL library [39] was also used to perform the geometric polygon computation, and the Python programming language was used for visualization. The SMP library [40] was used for RRT* algorithm implementation.

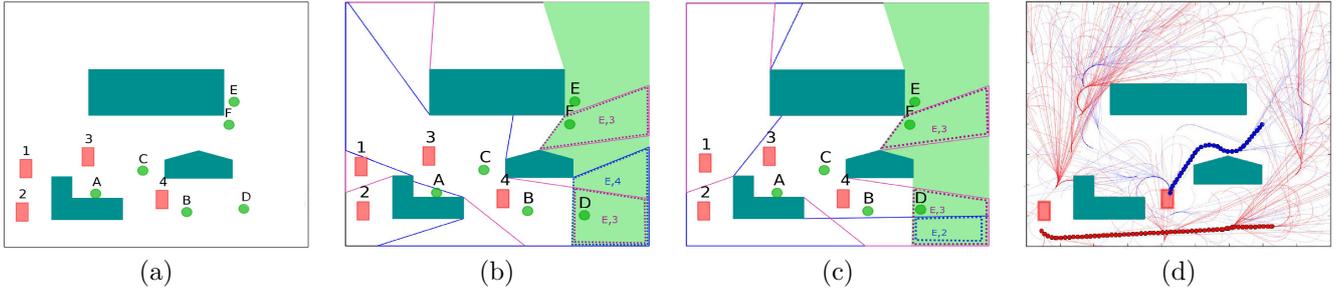

Figure 7: Bonnmotion random waypoint experiment: (a) Unit $E$ gets disconnected; (b) Goal region computation for candidate vehicle 2. The green shaded region is the visibility polygon of $E$. Purple dashed regions are the intersections of vehicle 3 and unit $E$'s visibility polygon while blue dashed area is the intersecting polygon of 4 and $E$. (c) Goal region for candidate vehicle 4. (d) RRT* trees and resulting trajectories for the two candidate vehicles 2 and 4.

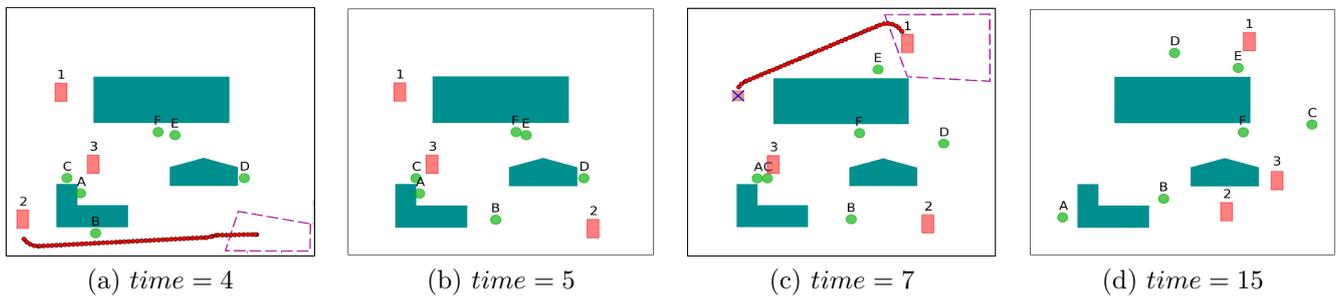

Figure 8: Bonnmotion random waypoint experiment at different times: (a) System reconnected by relocation of vehicle 2 to recover $D$. (b) System is still connected at $time = 5$. (c) Unit $E$ is disconnected and the system is recovered through relocation of vehicle 1. (d) An example system that is unrecoverable by a single vehicle movement.

In Figure 7, we present the test cases for a random waypoint mobility model where *four* servicing vehicles are assigned to monitor *six* deployed units. While the units are moving randomly, unit $E$ gets disconnected from the network. Therefore, we demonstrate the computation of our proposed Algorithm 3 in Figures 7(a)-(d) in order to recover the network. As the relocation of vehicle 1 or 3 would cause other vehicles to become disconnected from the network, they are both eliminated from consideration and the candidate vehicle set becomes $C = \{2, 4\}$. In Figure 7(b), we compute the goal region $X_G^2$ for vehicle 2. We have three regions to consider from the intersection of the visibility polygons marked by dotted lines. From among those, we select the region labeled as "$E, 4$" as $X_G^2$, which is the largest of the three. Similarly, we compute the region "$E, 3$" as the goal region $X_G^4$ for vehicle 4 shown in Figure 7(c). Finally, as shown in Figure 7(d), we generate two motion paths corresponding to the vehicles 2 (red) and 4 (blue) using the RRT* [41] motion planning algorithm for a Dubins car [26]. We select vehicle 4 for relocation by following the blue trajectory as it gives an optimal cost compared to the red trajectory which requires a longer path to travel.

In Figure 8(a) we have three available vehicles forming a relay network while serving six units. Unit $D$ gets disconnected and the candidate vehicle set for relocation is $C = \{2, 3\}$, as relocating vehicle 1 makes the relay network broken. As both of them have hard constrained units ($H_2 = \{B\}$; $H_3 = \{E, F\}$), we use (9) to calculate the intersecting polygons $X_G^2$ and $X_G^3$ as their respective goal regions. The resulting region that optimizes the visibility is shown as a dashed area for vehicle 2, which is the intersecting visibility region of vehicle 3, hard constrained unit $H_2 = \{B\}$, and disconnected unit $D$. Vehicle 2 is then relocated into the purple dashed region following the trajectory generated by the RRT* algorithm. At $time = 5$ as shown in Figure 8(b), the hard constrained unit $B$ of vehicle 2 changes position and does not break the connectivity. At $time = 7$ (Figure 8(c)) we dispatch vehicle 1 to serve the disconnected unit $E$ after the same computations done for the above cases. However, at $time = 15$ (Figure 8(d)), the system becomes non-recoverable when unit $A$ gets disconnected. We cannot move vehicle 1 or 2 because there is no common visibility polygon among hard constrained units, another vehicle, and the disconnected unit $A$. Vehicle 3 cannot be moved due to relay connectivity.

*Nomadic Mobility Model:* Units move in groups according to the nomadic mobility model and an example scenario is presented in Figure 9. We deployed two servicing vehicles in order to provide service to six units that are distributed into two groups (see Figure 9(a)). Unit $B$ gets disconnected in the next time-stamp shown in Figure 9(b). Accordingly, vehicle 1 goes to the intersection of the visibility polygons of vehicle 2 and unit $B$ (purple dashed).

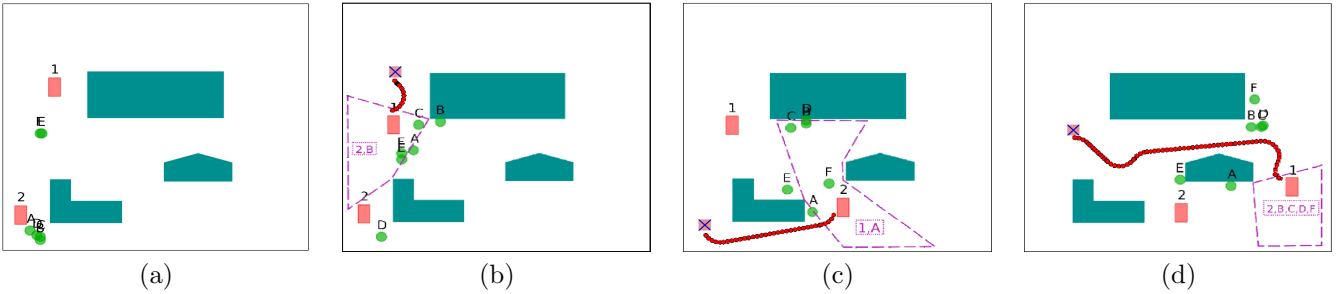

Figure 9: Bonnmotion nomadic mobility experiment: (a) All components are connected and the units form two groups; (b) Vehicle 1 is relocated to its goal region $X_G^1$ (purple area, which is the intersection of vehicle 2 and unit $B$) in order to serve disconnected unit $B$; (c) Vehicle 2 moves to serve disconnected unit $A$; (d) Again, vehicle 1 is dispatched to serve the disconnected unit $F$.

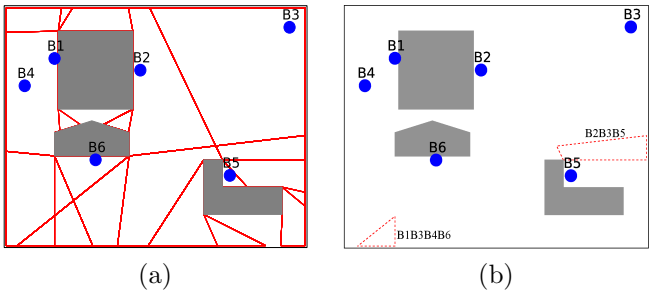

Figure 10: (a) Decomposition of an environment using visibility polygons. (b) Selected polygons using approximate set cover algorithm.

This small movement is highlighted by a red curvature generated by the RRT* algorithm. In Figure 9(c), unit $A$ is disconnected and vehicle 2 is dispatched to its calculated goal location $X_G^2$ (the intersection of the visibility from vehicle 1 and unit $A$). Then, in the next time-stamp, unit $F$ is disconnected. Only vehicle 1 has a common intersection with vehicle 2, hard constrained unit set $H_1 = \{B, C, D\}$ and disconnected unit $F$. Therefore, we relocate vehicle 1 to repair the LoS-based visibility network. We observed that the nomadic mobility model is easier to repair than the random waypoint model with a single vehicle movement as the units move in groups.

### 7.3. Re-Establishing a Communication-valid State

We have tested the methodology discussed in section 6.2 to establish a communication-valid network (static or dynamic) in case 1) a new setup is needed, or 2) there is no solution with a single vehicle movement once a connected network becomes communication-invalid. At first, a visibility-based polygonal decomposition of the environment was obtained using the VisiLibity [42] and Shapely [43] libraries as shown in Figure 10(a). After applying algorithm 4, we get the two polygons presented in Figure 10(b) that collectively see all the units. The bottom-left polygon is completely visible from units $B_1$, $B_3$, $B_4$ and $B_6$ while the units $B_2$, $B_3$ and $B_5$ can see the middle-right polygon.

We tested our approximate solution on several randomly generated environments with six ($m = 6$) units shown in Figure 11. Our algorithm was able to select the best polygons according to the labels $y_i$ and scores $\hat{s}_i$ in $\mathcal{F}$. Accordingly, a single vehicle can serve the units deployed in Figure 11(a) as we found a single polygon visible to all the units. In the case of 11(b), two vehicles are sufficient to cover the two selected polygons. Moreover, the polygons are completely visible to each other and form a single connected component. Therefore, no extra vehicle is required to do patrolling. Then, we need a minimum of three vehicles in the case shown in Figure 11(c), where two of them are assigned to each of the polygons while the remaining one connects the two polygons using an approximate TSP tour $\tau$. A scenario is presented in Figure 11(d) with three selected polygons. We need three vehicles to form a static relay network as the three polygons are completely visible to each other and therefore no further patrolling is required. However, patrolling needs to be planned in case we have less than three available vehicles.

An animated simulation model is developed using ROS and the Gazebo 3D simulator [1] where we use a number of Husky cars as our robot vehicles as shown in Figure 12. The Husky is a simulated version of *Clearpath Robotics* real UGV, and is widely used in research for its enabling of realistic simulation of real world hardware capabilities. The six cylindrical objects in Figure 12(a) represent the six ($m = 6$) units, while the black vehicle around the center is the Husky UGV. We first decompose the environment based on visibility polygons of the units as shown in Figure 12(b) and three polygons $P_1$, $P_2$ and $P_3$ are selected by the approximate set cover method of Algorithm 4 (see Figure 12(c)).

Next, we simulated two cases with one ($n = 1$) and three ($n = 3$) *Husky* cars as shown in Figure 12(c)-(d). Given a single vehicle, a motion planning algorithm generates a tour $\tau$ that touches the chosen polygons, avoiding the obstacles (Figure 12(c)). The visiting order of the polygons is obtained from the approximate TSP [37] algorithm as discussed in Section 6.3. The Husky vehicle uses a motion planner that combines A* with Adaptive Monte Carlo Localization (AMCL) [44] to move the car from source to goal. However, one can use any other motion planner that conforms the robots dynamics and configuration, such as RRT* or PRM*. In Figure 12(d), a case is presented with three available vehicles, so we can assign one vehicle per polygon which serves as static servers. As a result, we get a connected component $P_{C_1}$ and a uncovered polygon $P_3$ as explained in Section 6.3. Therefore, the remaining vehicle follows a patrolling trajectory $\tau$ in between $P_{C_1}$ and $P_3$. Detailed simulations with animation can be found in the attached multimedia of this paper.

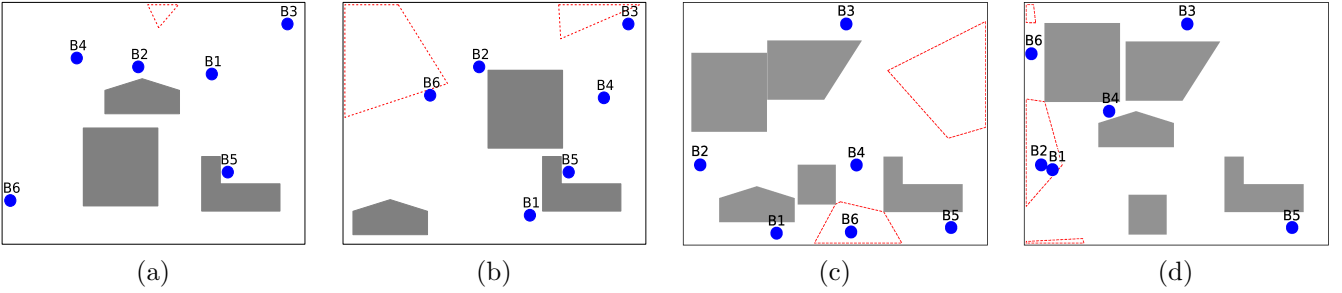

Figure 11: (a) A single vehicle is sufficient to serve all the units as $|\Gamma| = 1$; (b) Two vehicles are sufficient as their deployment will result in a single connected component; (c) Three vehicles are required where two of them will be deployed in two goal polygons and the remaining one will do the patrolling between their visibility polygons; (d) Three vehicles can form a static relay network as the three goal polygons are completely visible to each other.

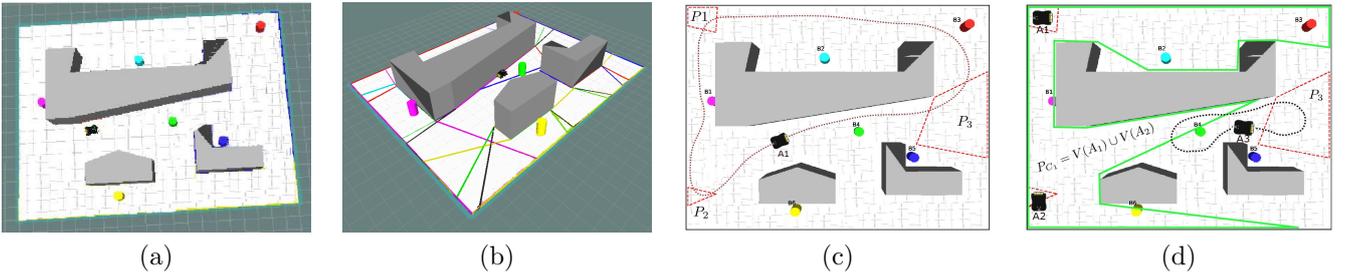

Figure 12: (a) ROS and Gazebo simulation environment containing six units presented with various colors; (b) Visibility based decomposition; (c) Planning with one vehicle; (d) Planning with three available vehicles.

*7.4. Physical Deployment*

We performed a physical experiment of our ideas using a modified Traxxas Slash Dakar Truck Series Edition as a servicing vehicle as shown in Figure 13(a). An *ArduPilot* controller (ArduPilot Mega APM 2.5) was added to control the movement of the vehicle. A Turnigy 9X radio was used to place a series of waypoints to be followed by the vehicle and as a safety feature in case of communication loss. The Raspberry Pi (version 2) unit was mounted for on-board processing and an external compass/GPS unit was mounted for localization.

We connected a camera (Vilros 5MP Camera Board Module) to the Raspberry Pi as a visibility sensor and used simple color segmentation algorithms for unit detection using the *OpenCV* computer vision library. On top of each unit, a Zigbee communication module (Zigbee+Arduino) was used to communicate with the vehicle as shown in Fig-ure 13(b). In Figure 13(c) and (d) we see the output of unit detection captured by the on-board camera mounted on the vehicle when the units come within the visibility region of the vehicle. The images are processed in real time using onboard processing power while the vehicle is in motion. The Zigbee module is used for passing messages between the vehicle and the units. Detailed experiments with the robots in action can be found in the attached multimedia of this paper.

## 8. Conclusions and Future Work

In this paper, we study the problem of establishing Line-of-Sight (LoS) communication between a number of moving vehicles and a group of mobile units. We proposed algorithms to determine if a configuration of units and vehicles is connected through LoS communication. Two polynomial-time versions of the algorithm, one centralized and one distributed, were developed to be deployed for different types of ground missions. Secondly, we proposed a complete algorithm that gives a solution, if there is any, to recover a system by relocating a single vehicle.

In a complex and highly dynamic situation, where a single vehicle fails to repair the network, we solve the general problem of multi-robot relocation and placement. We proved that the exact solution to this problem is NP-hard and then presented heuristic procedures based on set cover approximation to calculate goal locations and paths. In a patrolling scenario with insufficient vehicles, we use the TSP algorithm to visit the calculated goal polygons. Further optimization was achieved in terms of motion or patrolling cost through visibility-based geometry and graph algorithms. Finally, the ideas were extensively tested in a realistic simulated environment with the help of ROS, Gazebo, and the simulated Husky UGV, and in an outdoor experimental deployment with a modified RC car. Several interesting directions are left for future work.

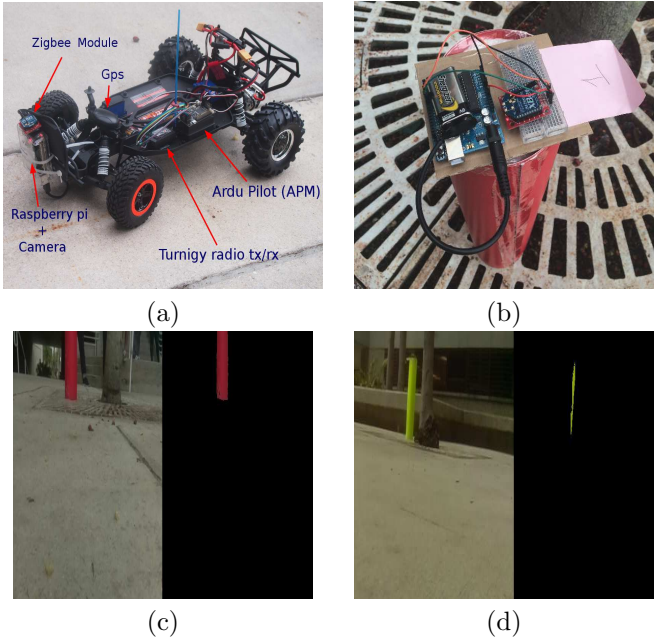

Figure 13: (a) Modified robotic truck as a servicing vehicle with APM, Raspberry pi, GPS, Zigbee and Camera mounted on it; (b) A sample unit (Red) with a Zigbee module mounted on it as a communication device; (c) and (d) are the images captured by the camera mounted on the vehicle along with their real time Computer vision output after color based segmentation (to detect red and yellow) shown on the right side of each image.

We found that the problem of interest is NP-hard and can only be approximated with an $O(\log n)$ ratio at best. We presented a heuristic solution inspired by set cover approximation that uses visibility polygon decomposition as input and TSP with neighbors for a patrolling sequence. It is clear that this finds a feasible solution, but calculating the exact approximation ratio is still an open problem.

Another extension of our work is to remove assumptions about the known world, $\mathcal{W}$, and obstacles, $\mathcal{O}$. We assumed that the obstacles are known beforehand and the layout can be perfectly decomposed. Ideally, a robot equipped with sensors can create a strategy based on visibility events [25] to explore the environment and find good LoS locations. We are exploring the related problem of finding competitive strategies for a kernel polygon search and determining if they can be implemented in a mobile vehicle with sensors [45]. We also want to remove the need to estimate the state of the units to follow the proposed path. A feedback based planning approach using the function $\hat{s}$ as a navigation function may help to overcome this problem.

## References

[1] N. Koenig, A. Howard, Design and use paradigms for gazebo, an open-source multi-robot simulator, in: Intelligent Robots and Systems, 2004.(IROS 2004). Proceedings. 2004 IEEE/RSJ International Conference on, Vol. 3, IEEE, 2004, pp. 2149–2154.

[2] M. M. Rahman, L. Bobadilla, B. Rapp, Establishing line-of-sight communication via autonomous relay vehicles, 2016 IEEE Military Communications Conference (MILCOM).

[3] A. Dumitrescu, J. S. Mitchell, Approximation algorithms for tsp with neighborhoods in the plane, in: Proceedings of the twelfth annual ACM-SIAM symposium on Discrete algorithms, Society for Industrial and Applied Mathematics, 2001, pp. 38–46.

[4] J. S. Mitchell, Geometric shortest paths and network optimization, Handbook of computational geometry 334 (2000) 633–702.

[5] J. O'Rourke, Art Gallery Theorems and Algorithms, Oxford University Press, New York, 1987.

[6] J. O'Rourke, Visibility, in: J. E. Goodman, J. O'Rourke (Eds.), Handbook of Discrete and Computational Geometry, 2nd Ed., Chapman and Hall/CRC Press, New York, 2004, pp. 643–663.

[7] D. Kirkpatrick, Optimal search in planar subdivisions, SIAM Journal on Computing 12 (1) (1983) 28–35.

[8] H. González-Baños, J.-C. Latombe, A randomized art-gallery algorithm for sensor placement, in: Proceedings of the seventeenth annual symposium on Computational geometry, ACM, 2001, pp. 232–240.

[9] L. Erickson, S. M. LaValle, An art gallery approach to ensuring that landmarks are distinguishable, in: Proceedings Robotics: Science and Systems, 2011.

[10] J. S. Mitchell, Approximating watchman routes, in: Proceedings of the Twenty-Fourth Annual ACM-SIAM Symposium on Discrete Algorithms, SIAM, 2013, pp. 844–855.

[11] K. J. Obermeyer, P. Oberlin, S. Darbha, Sampling-based path planning for a visual reconnaissance unmanned air vehicle, Journal of Guidance, Control, and Dynamics 35 (2) (2012) 619–631.

[12] L. J. Guibas, J.-C. Latombe, S. M. LaValle, D. Lin, R. Motwani, Visibility-based pursuit-evasion in a polygonal environment, in: F. Dehne, A. Rau-Chaplin, J.-R. Sack, R. Tamassia (Eds.), WADS '97 Algorithms and Data Structures (Lecture Notes in Computer Science, 1272), Springer-Verlag, Berlin, 1997, pp. 17–30.

[13] S. Bhattacharya, R. Murrieta-Cid, S. Hutchinson, Optimal paths for landmark-based navigation by differential-drive vehicles with field-of-view constraints, Robotics, IEEE Transactions on 23 (1) (2007) 47–59.

[14] T. Muppirala, R. Murrieta-Cid, S. Hutchinson, Optimal motion strategies based on critical events to maintain visibility of a moving target, in: Proceedings IEEE International Conference on Robotics & Automation, 2005, pp. 3837–3842.

[15] D. A. Anisi, P. Ogren, X. Hu, Cooperative minimum time surveillance with multiple ground vehicles, IEEE Transactions on Automatic Control 55 (12) (2010) 2679–2691.

[16] J. Banfi, A. Q. Li, N. Basilico, I. Rekleitis, F. Amigoni, Asynchronous multirobot exploration under recurrent connectivity constraints, in: IEEE International Conference on Robotics and Automation (ICRA), 2016.

[17] Y. Kantaros, M. M. Zavlanos, Distributed intermittent connectivity control of mobile robot networks, IEEE Transactions on Automatic Control 62 (7) (2017) 3109–3121.

[18] G. A. Hollinger, S. Singh, Multirobot coordination with periodic connectivity: Theory and experiments, IEEE Transactions on Robotics 28 (4) (2012) 967–973.

[19] A. Monfared, M. Ammar, E. Zegura, D. Doria, D. Bruno, Computational ferrying: Challenges in deploying a mobile high performance computer, in: World of Wireless, Mobile and Mul-

timedia Networks (WoWMoM), 2015 IEEE 16th International Symposium on a, IEEE, 2015, pp. 1–6.
[20] D. Bhadauria, O. Tekdas, V. Isler, Robotic data mules for collecting data over sparse sensor fields, Journal of Field Robotics 28 (3) (2011) 388–404.
[21] M. Dunbabin, P. Corke, D. I. Vasilescu Rus, Data muling over underwater wireless sensor networks using an autonomous underwater vehicle, in: Robotics and Automation, 2006. ICRA 2006. Proceedings 2006 IEEE International Conference on, IEEE, 2006, pp. 2091–2098.
[22] O. Tekdas, V. Isler, J. Lim, A. Terzis, Using mobile robots to harvest data from sensor fields, IEEE Wireless Communications 16 (1) (2009) 22.
[23] J. C. Juarez, A. Dwivedi, A. R. Hammons, S. D. Jones, V. Weerackody, R. A. Nichols, Free-space optical communications for next-generation military networks, Communications Magazine, IEEE 44 (11) (2006) 46–51.
[24] M. Khan, M. Yuksel, Maintaining a free-space-optical communication link between two autonomous mobiles, in: Wireless Communications and Networking Conference (WCNC), 2014 IEEE, IEEE, 2014, pp. 3154–3159.
[25] S. M. LaValle, Planning algorithms, Cambridge university press, 2006.
[26] J. T. Isaacs, D. J. Klein, J. P. Hespanha, Algorithms for the traveling salesman problem with neighborhoods involving a dubins vehicle, in: American Control Conference (ACC), 2011, IEEE, 2011, pp. 1704–1709.
[27] H. El Gindy, D. Avis, A linear algorithm for computing the visibility polygon from a point, Journal of Algorithms 2 (2) (1981) 186–197.
[28] M. M. Zavlanos, M. B. Egerstedt, G. J. Pappas, Graph-theoretic connectivity control of mobile robot networks, Proceedings of the IEEE 99 (9) (2011) 1525–1540.
[29] E. Stump, A. Jadbabaie, V. Kumar, Connectivity management in mobile robot teams, in: Robotics and Automation, 2008. ICRA 2008. IEEE International Conference on, IEEE, 2008, pp. 1525–1530.
[30] R. Merris, Laplacian matrices of graphs: a survey, Linear algebra and its applications 197 (1994) 143–176.
[31] S. Bhattacharya, T. Başar, Graph-theoretic approach for connectivity maintenance in mobile networks in the presence of a jammer, in: Decision and Control (CDC), 2010 49th IEEE Conference on, IEEE, 2010, pp. 3560–3565.
[32] S. Karaman, M. R. Walter, A. Perez, E. Frazzoli, S. Teller, Anytime motion planning using the rrt*, in: Robotics and Automation (ICRA), 2011 IEEE International Conference on, IEEE, 2011, pp. 1478–1483.
[33] S. Har-Peled, Geometric approximation algorithms (Chapter 17), Vol. 173, American mathematical society Providence, 2011.
[34] H. Brönnimann, M. T. Goodrich, Almost optimal set covers in finite vc-dimension, Discrete & Computational Geometry 14 (4) (1995) 463–479.
[35] T. H. Cormen, Introduction to algorithms, MIT press, 2009.
[36] J. Hopcroft, R. Tarjan, Algorithm 447: efficient algorithms for graph manipulation, Communications of the ACM 16 (6) (1973) 372–378.
[37] N. Christofides, Worst-case analysis of a new heuristic for the travelling salesman problem, Tech. rep., DTIC Document (1976).
[38] N. Aschenbruck, R. Ernst, E. Gerhards-Padilla, M. Schwamborn, Bonnmotion: a mobility scenario generation and analysis tool, in: Proceedings of the 3rd International ICST Conference on Simulation Tools and Techniques, ICST (Institute for Computer Sciences, Social-Informatics and Telecommunications Engineering), 2010, p. 51.
[39] A. Fabri, S. Pion, Cgal: The computational geometry algorithms library, in: Proceedings of the 17th ACM SIGSPATIAL international conference on advances in geographic information systems, ACM, 2009, pp. 538–539.
[40] S. Karaman, E. Frazzoli, Sampling-based motion planning (smp) library, in: https://svn.csail.mit.edu/smp, MIT.
[41] S. Karaman, M. R. Walter, A. Perez, E. Frazzoli, S. Teller, Anytime motion planning using the rrt*, in: Robotics and Automation (ICRA), 2011 IEEE International Conference on, IEEE, 2011, pp. 1478–1483.
[42] K. J. Obermeyer, Contributors, The VisiLibity library, http://www.VisiLibity.org, r-1 (2008).
[43] S. Gillies, A. Bierbaum, K. Lautaportti, O. Tonnhofer, Shapely, URL: http://toblerity.org/shapely.
[44] D. Fox, W. Burgard, F. Dellaert, S. Thrun, Monte carlo localization: Efficient position estimation for mobile robots, Association for the Advancement of Artificial Intelligence AAAI.
[45] C. Icking, R. Klein, Searching for the kernel of a polygon- a competitive strategy, in: Proceedings of the eleventh annual symposium on Computational geometry, ACM, 1995, pp. 258–266.